\pdfoutput=1

\documentclass[paper=a4,fontsize=10pt,DIV=12]{scrartcl}

\usepackage[english]{babel} 

\usepackage[latin1]{inputenc}
\usepackage[T1]{fontenc}
\usepackage{lmodern}
\typearea[current]{last}

\usepackage{color}

\usepackage[activate,final]{microtype}
\usepackage{xspace}

\usepackage{etoolbox}

\usepackage{enumitem}
\setlist{nolistsep}
\usepackage[english=british,german=guillemets]{csquotes}
\usepackage{booktabs}\usepackage{flafter}
\usepackage{bm}

\usepackage{amsmath,amssymb,amsthm}
\usepackage{mathtools}

\usepackage{cases}
 \newtheorem{theorem}{Theorem}[section]
 \newtheorem{lemma}[theorem]{Lemma}

 \theoremstyle{definition}
 \newtheorem{definition}[theorem]{Definition}

 \usepackage{tikz}
\usetikzlibrary{calc,arrows,decorations,decorations.pathreplacing,backgrounds,shapes,fit}

\addtokomafont{sectioning}{\rmfamily}
\addtokomafont{descriptionlabel}{\rmfamily}
\makeatletter
\patchcmd{\subject@font}{\Large}{\large}{}{}
\patchcmd{\@maketitle}{\Large}{\large}{}{}
\patchcmd{\@maketitle}{\Large}{\large}{}{}
\patchcmd{\@maketitle}{\Large}{\large}{}{}
\patchcmd{\@maketitle}{\Large}{\large}{}{}
\patchcmd{\@maketitle}{\titlefont\huge}{\titlefont\LARGE}{}{}
\makeatother

\DeclarePairedDelimiter{\p}{\lparen}{\rparen}

\DeclarePairedDelimiter{\cb}{\{}{\}}
\DeclarePairedDelimiter{\br}{[}{]}

\newcommand{\eq}[2][]{\begin{equation}\label{#1}\begin{split}#2\end{split}\end{equation}}

\let\phi\varphi

\newcommand{\cnf}{\textsc{\mdseries CNF}}

\usepackage{colonequals}

\usepackage[final,
            pdfusetitle,
            bookmarksopen=true,
            pdfstartview=FitH,
            pdfprintscaling=None,
            unicode=true,
            ]{hyperref}

\usepackage{algorithm2e}            
            \title{AFP Algorithm and a Canonical Normal Form for Horn Formulas 
}
\author{Ruhollah Majdoddin\\
  Humboldt Universität zu Berlin, Germany\\\texttt{r.majdodin@gmail.com}}
\date{}

\begin{document}
\SetKwComment{cmt}{*}{}   
\maketitle

\begin{abstract} 
AFP Algorithm is a learning algorithm for Horn formulas. We show that it does not improve the complexity of  AFP Algorithm, if after
each negative counterexample more that just one refinements  are performed. Moreover,  a
canonical normal form for  Horn formulas is presented, and it is proved that the
output formula of AFP Algorithm is in this normal form.
\end{abstract}
\begin{section}{Introduction}
In propositional logic, a \textit{literal} is a variable or a negated variable. A \textit{clause}
is  a disjunction of literals. A \textit{Horn clause} is a clause with at most one
   unnegated variable. A \textit{Horn formula} is a conjunction of Horn
  clauses. Clearly, the the class of Horn formulas is a subclass of \cnf.

  Angluin et al. \cite{AFP92} presented the AFP Algorithm, which is a polynomial
  time \textit{learning algorithm} for Horn
  formulas. This means that for a certain Horn formula (unknown to the
  algorithm), the algorithm makes some membership queries and
  outputs an equivalent Horn formula.
  
Arias et al. \cite{AB11} presented a \textit{normal form} for Horn formulas,
which means every Horn formula can be converted to an equivalent formula in this
normal form. Moreover, they proved  that the normal form is \textit{canonical}, that is every Horn formula has only
one equivalent formula in this normal form,
up to the order of clauses. 
They presented an algorithm that given a Horn formula, outputs its
canonical normal formula. Moreover, they proved that the output formula of the
AFP Algorithm is in this canonical normal form. 

We also have  independently discovered this canonical normal form and in section 2 we prove that the output of 
the AFP Algorithm is in this form. Our presentation is briefer than
\cite{AB11}.

Balc{\'a}zar \cite[section 7]{Bal05} poses as a frequently asked but unanswered question, that
whether it improves the time and query complexities of AFP Algorithm, if we change the algorithm, so that after every negative
counterexample, it makes more that just one refinements. Interestingly, \cite{AFP92} had
already briefly answered this question, negatively. Here we provide a detailed proof for that. 

%

\begin{subsection}{Preliminaries}
We follow the notation of \cite{AB11}. Additionally we denote the variables with
letters from the beginning of the alphabet, and the set of variables of a
formula with
\(V\). Logical False will be denoted by \(F\).  Subset and proper subset are denoted by \(\subseteq\) and \(\subset\) respectively. With
a formula  we mean a Horn formula. With \(a^{n}\) where \(a \in \cb{0,
1}\), we mean a string of \(a\)'s of length \(n\).
%
\end{subsection}
\end{section}
\begin{section}{Canonical normal form}
We first define our normal form and then prove that it is indeed a normal form.
\begin{definition}[Normal Form] 
\label{canonical-normal-form}
A formula \(H = \bigwedge_i \p{\alpha_i \rightarrow \beta_i}\) is in \emph{normal form}, if 

\begin{enumerate}
  \item \(\alpha_i \neq \alpha_j\) for  \(i \neq j\),
   \item \(\alpha_i \subset \beta_i\),
  \item \(\forall i,j \,\, \alpha_j \vDash \p{\alpha_i \rightarrow \beta_i}\)\label{consistency}.
\end{enumerate}
\end{definition}
Compare this Definition with \cite[Definition 4 in Section 3]{AB11}. 
\begin{theorem}\label{}
Each formula has an equivalent formula in normal form. 
\end{theorem}
\begin{proof}
 We present a polynomial time algorithm that given a formula, outputs an
 equivalent formula in  normal form

Repeat until no more changes are made:
\renewcommand{\labelenumi}{\alph{enumi})}
\begin{enumerate}
  \item Merge the clauses with the same antecedents.
  \item For each clause \(\alpha \rightarrow \beta\), if there is a clause \(\kappa \rightarrow \gamma\) such
  that \(\kappa \subset \alpha\), then replace \(\alpha \rightarrow \beta\) with \(\alpha \cup
  \gamma\rightarrow \beta\)  
\end{enumerate}
At the end,

\noindent 
c) Delete all clauses \(\alpha \rightarrow \beta\) that \(\beta \subseteq \alpha \).

\noindent
d) Change all clauses \(\alpha \rightarrow \beta\) to \(\alpha \rightarrow \alpha \cup \beta\).
 
As there are finite variables, and at each iteration only some clauses are merged or the size of some
antecedents increases, the iteration (and algorithm) end in polynomial time. Properties 1, 2, and 3 of the
normal form will be fulfilled by a, (c and d) and b respectively.  
\end{proof}
AFP Algorithm \cite{AFP92} uses a learning protocol to make membership queries
to a certain formula (the formula is not explicitly given), and outputs an
equivalent formula. In the following, we give a slightly different version of
AFP Algorithm. 
%
\begin{algorithm}
\DontPrintSemicolon
\textbf{AFP}\;   
\(O \leftarrow \p{}\)   \cmt*[f] {List of sets}\;
\(P \leftarrow \cb{}\)  \cmt*[f] {Set of positive counterexamples}\;
\(H \leftarrow T\)  \cmt*[f] {Set the \(H\) equal to True.}\;
\While {\(\text{equal}\p{H, H^*} = \p{"no", y}\)}{ 

\If{\(y \nvDash H\)}{\cmt*[f]{a positive counterexample} \;
\(P \leftarrow P \cup \cb{y} \) 
}
\Else{
\cmt*[f]{a negativ counterexample} \;
\nl \For {the first \(s \in O\), such that \(\textit{member}\p{s \wedge y} = "no"\) and \(\br{s \wedge y} \subset
\br{s} \)}
{\nllabel{bestandteil_anfrage} 
\(s \leftarrow s \wedge y\)	\nllabel{verfeinerung}
}
	\If{none is found}{
		Add \(y\) as the last element in \(O\) \nllabel{neue_menge} 
	}	
}
 Set \(H\) as the conjunction of all \(\cb{s \rightarrow a \;\big|\; s \in O, \; a \in V \cup
 \cb{F}\text{ And } \forall z \in P \;\; z \vDash s \rightarrow a}\)\nllabel{alle_konklusionen} 
} 
output \(H\)\;
\caption{the AFP Algorithm}
\end{algorithm}
%

In the rest of this section, we shall prove that the normal form is canonical,
and that the output formula of AFP  Algorithm is
in normal form.

\begin{lemma}\label{negativ-ce-theorem}
Let \(H^* = \bigwedge_i \p{\alpha_i \rightarrow \beta_i}\) be a normal form
formula equivalent to the target formula for
 AFP Algorithm. While there are still antecedents of  \(H^*\) which are equal to no antecedent of
\(H\) (equivalently to no  \(s \in O\)), there will be a negative counterexample.

\end{lemma}
\begin{proof}  
Let \(\alpha\) be such an antecedent, then \(\alpha \nvDash H^*\). If there are no clauses \(s \rightarrow
\gamma\) from \(H\), that \(s \subset \alpha\), then \(\alpha \vDash H\) as a negative Counterexample. Else,
on line \ref{alle_konklusionen}, \(\gamma \supseteq \bigcup_{i, \alpha_i \subseteq s}
\beta_i \setminus s\), and there are finite positive counterexamples until the equality for all such \(s\)
holds. And then, following the definition of normal form
(definition \ref{canonical-normal-form}-\ref{consistency}), \(\gamma \subset \alpha\), so \(\alpha \vDash s
\rightarrow \gamma\) and \(\alpha \vDash H\) will be the negative counterexample.
\end{proof}

Let \(H^* = \bigwedge_i \p{\alpha_i \rightarrow \beta_i}\) be a 
normal form formula equivalent to the target formula. Then the list  \(O=\p{s_i}\) in the AFP Algorithm, has
the \emph{property} that 
\eq{
\text{At each instant, } \ \forall i	\exists  k \quad  s_i \nvDash \alpha_k
\rightarrow \beta_k \text{ And }   \p{s_j \nvDash \alpha_k
\rightarrow \beta_k	\Rightarrow  j \geq i}  }  

\begin{lemma}[Property of \(O\)]
\label{property-of-AFP}  
The above property holds.
\end{lemma} 
Compare this with \cite{AB11}, Lemma 16. 
\begin{proof}
When an \(s_i \in O\) is added or after it is refined, it holds that  \(s_i\) violates at least one
clause \(\alpha \rightarrow \beta\) from \(H^*\), such that \(\forall j<i \quad \alpha \nsubseteq s_j\). That holds at least
untill the next time that \(s_i\) is refined.
\end{proof}

\begin{theorem}
\label{afp-funktioniert}
The AFP Algorithm returns a  formula in normal form that is
equivalent to the the target formula.
\end{theorem}
\begin{proof}
Let \(H^*\) be a formula in normal form equivalent to the target formula. Via each negative counterexample, some members of a
set in \(O\) are removed (Line \ref{verfeinerung}), or a new set is added to \(O\) (Line \ref{neue_menge}). Theorem \ref{property-of-AFP} implies that the
size of \(O\) is no more than the number of clauses of \(H^*\). Following Theorem \ref{negativ-ce-theorem},
while there are antecedents of \(H^*\) that are not identical with some set in \(O\), negative counterexamples will be given. As it is all finite, \(O\) will be equal to set of the
antecedents of \(H^*\) and the conclusions will corrected via positive counterexamples. So the algorithm
ends, and it then holds that \(H = H^*\).
\end{proof}
\begin{theorem}[Canonicality]
Each formula has exactly one equivalent formula in normal form, up to the order of clauses.
\end{theorem}
\begin{proof}
It follows from the free choice of \(H^*\) in proof of theorem \ref{afp-funktioniert}.
\end{proof} 
\end{section} 
\begin{section}{More than one refinement with each negative counterexample}
It seems an appealing question, that whether AFP Algorithm would be more efficient, in
runtime and number of queries, had it tried to refine more that one set of \(O\) with each negative
counterexample, so far that \cite[section 7]{Bal05} considers it as a frequently asked unanswered
question.
Interestingly \cite{AFP92} had already briefly answered this question:
``Overzealous refinement may result in  several  examples  in \(O\)  violating 
the  same  clause  of \(H^*\).  To  avoid  this,  whenever  a  new  negative  counterexample  could  be 
used  to  refine  several  examples  in  the  sequence  \(O\),  only  the  first  among  these  is  refined.''
Here we provide a proof for this answer.

Besides AFP Algorithm , Angluin et al. \cite{AFP92} present a second
version of AFP Algorithm, (which they call Horn1),  
which is more efficient in determining the conclusions, but makes the same as AFP in finding the
antecedents. We will show that the answer to the above question (for worst-case
runtime) is negative for both algorithms. Let the algorithm AFP* be the same as AFP
Algorithm, but ``the first'' in line \ref{bestandteil_anfrage} be replaced by
``for all''.

Throughout this section, let \(H^* = \bigwedge_i \p{\alpha_i \rightarrow
  \beta_i}\) be the formula in canonical normal
form that is equivalent to the target formula, such that antecedents with smaller size, have smaller index. We say that the
sequence of counterexamples in a run of the AFP Algorithm is \emph{ordered} (relative to \(H^*\)) if it is
as follows.

The sequence of counterexamples is a succession of \(m\)
subsequences\footnote{Here each subsequence simply consists of consecutive parts
of the sequence.}, where \(m\) is the number of clauses of \(H^*\).
Let \(\p{z_j}\) be the \(i\)th subsequence. Then each \(z_j\) is a superset of  \(\alpha_i\) but not a
superset of \(\beta_i\) and not a superset of any \(\alpha_k\) such that \(\alpha_k\not\subset \alpha_i\); Moreover
 \(z_{j+1} \cap z_{j} \subset z_j \), and the last element of \(\p{z_j}\) is \(\alpha_i\).
 Note that in this definition there is no restriction on positive counterexamples or their order relative to
 negative counterexamples.

It is straightforward to show that for every target formula, there exists at least one ordered sequence of
couterexaples. An example will be given in the proof of theorem \ref{AFP*-nobetterthan-AFP}.

%
 
\begin{lemma}
\label{AFP*-AFP-least-helpful}
For any Horn formula, if the (negative) counterexamples are ordered,
then the AFP and AFP* algorithms perform exactly the same operations.
\end{lemma}
\begin{proof}
By induction we show that after the \(i\)th subsequence, \(s_i = \alpha_i\) and
it will not be changed later. So let it be true for the first \(i-1\)
subsequences.
Then at round \(i\), for any of the negative counterexamples \(z_k\) if the
test of line \ref{bestandteil_anfrage} is true for \(s_j = \alpha_j,\; j<i\), then it should be refined, but then as the index of clauses
of \(H^*\) is ordered by the size of antecedents, Property of \(O\) (theorem \ref{property-of-AFP}) cannot be satisfied.

Therefore during round \(i\) only \(s_{i}\) can be added or refined and as
the last negative counterexample in round \(i\) is \(\alpha_i\), with a similar
argument, \(s_i\) will be refined to \(\alpha_i\).
\end{proof}

\begin{theorem}
\label{AFP*-nobetterthan-AFP}
The worst-case time, equivalence and membership query complexities of the AFP* algorithm is not better than
that of AFP Algorithm.
\end{theorem}
\begin{proof}
For the target formulas from the class \(\cb{\phi_n}\) defined below, if the counterexamples are given as
described below, by lemma \ref{AFP*-AFP-least-helpful}, AFP* Algorithm and AFP Algorithm perform exactly
the same operations, because the (negative) counterexamples are ordered. But 
AFP Algorithm with this setting will reach its worst-case time, equivalence and
membership query complexities on general input (Compare with \cite[Theorem 2]{AFP92}, we do not repeat that argument here). The result follows.

The formula \(\phi_n\) in canonical normal form is defined with the  set of \(2n+1\) literals \(V=\cb{a_1, \ldots, a_{2n+1}}\),
and  has \(m=n\) clauses.
\[
\phi_n = \bigwedge_{1 \le i \le n} \p{a_i \rightarrow a_{2n+1}} 
\] 
The \(i\)th subsequence of negative counterexamples are
\eq{
y_1 &= 0^{i-1}\,1\,0^{n-i}\,1^{n}\,0\\ 
y_j &= 0^{i-1}\,1\,0^{n-i}\,1^{n-j+1}\,0^{j-1}\,0\\
y_{n+1} &= 0^{i-1}\,1\,0^{n-i}\,0^{n}\,0\\
}
After any \(y_j\) a sequence of positive counterexamples \(\p{w_k}\) will be given:
\eq{
w_k = V - \cb{d_k}
}
where \(d_k\) is the \(k\)th variable such that \(d_k \notin y_j\) and \(d_k \ne a_{2n+1}\).
\end{proof}
It is straightforward to give classes of formulas and counterexamples for which AFP Algorithm has its
general worse-case runtime and query complexities while algorithm AFP* makes substantially worse (roughly speaking,
because property of \(O\) (theorem \ref{property-of-AFP}) no more holds).
But we find theorem \ref{AFP*-nobetterthan-AFP} enough and more interesting. 
By a similar argument one can get similar results for algorithm Horn1.
\end{section}

 \bibliography{kuk}
\end{document}